\documentclass{llncs}
\usepackage{makeidx}
\usepackage{cite}
\usepackage{graphicx}
\usepackage{subfigure}
\usepackage[cmex10]{amsmath}
\usepackage{algorithmic}
\usepackage{url}
\usepackage{enumitem}
\usepackage{fancyhdr}
\fancypagestyle{ConferenceName}{\fancyhf{}\fancyfoot[C]{\footnotesize{Presented as a paper at the Workshop on Parallel and Distributed Computing for Knowledge Discovery in Databases (PDCKDD 2015) as a part of European Conference on Machine Learning and Principles and Practices of Knowledge Discovery in Databases (ECML/PKDD 2015)}}}
\begin{document}

\title{Faster learning of deep stacked autoencoders on multi-core systems using synchronized layer-wise pre-training}
\titlerunning{Synchronized layer-wise pre-training}

\author{Anirban~Santara\inst{1}, Debapriya~Maji\inst{2}, DP~Tejas\inst{1}, Pabitra~Mitra\inst{1} and Arobinda~Gupta\inst{1}} 
\authorrunning{Santara et al.} 

\institute{Department of Computer Science and Engineering, Indian Institute of Technology Kharagpur, Kharagpur 721302, WB, India
\and
Department of Electronics and Electrical Communication Engineering, Indian Institute of Technology Kharagpur, Kharagpur 721302, WB, India\\
$\mathtt{anirban\_santara@iitkgp.ac.in}$
}

\maketitle
\thispagestyle{ConferenceName}

\begin{abstract}
Deep neural networks are capable of modelling highly non-linear functions by capturing different levels of abstraction of data hierarchically. While training deep networks, first the system is initialized near a good optimum by greedy layer-wise unsupervised pre-training. However, with burgeoning data and increasing dimensions of the architecture, the time complexity of this approach becomes enormous. Also, greedy pre-training of the layers often turns detrimental by over-training a layer causing it to lose harmony with the rest of the network. In this paper a synchronized parallel algorithm for pre-training deep networks on multi-core machines has been proposed. Different layers are trained by parallel threads running on different cores with regular synchronization. Thus the pre-training process becomes faster and chances of over-training are reduced. This is experimentally validated using a stacked autoencoder for dimensionality reduction of MNIST handwritten digit database. The proposed algorithm achieved 26\% speed-up compared to greedy layer-wise pre-training for achieving the same reconstruction accuracy substantiating its potential as an alternative.
\end{abstract}


\section{Introduction}
\label{intro}
Artificial neural networks that have more than three hidden layers of neurons are called Deep Neural Networks (DNN)\cite{Erhan2010}. Every layer of a trained DNN represents a higher level abstraction of the input data than the previous one. DNNs are highly successful and provide the current state-of-the art results in many AI tasks. Stacked autoencoder is a class of DNN that is used for unsupervised representation learning and dimensionality reduction \cite{Bengio2009}. However, training deep architectures faced challenges ever since their introduction. The cost function being non-linear, gradient based routines usually get stuck in poor local minima for large initialization weights. On the other hand, with small initial weights backpropagated error gradients in the layers close to the input become tiny resulting in no appreciable training of those layers \cite{Hinton2006}. Greedy layer-wise pre-training was introduced with a motivation to approximately capture statistical abstractions of the input data in every layer and initialize the network parameters in a region near a good local optimum \cite{Hinton2006a,Bengio2006}. This was followed by fine-tuning via backpropagation to push the solution deeper into the optimum. This approach of training deep architectures has brought great success in many areas of application \cite{Bengio2009,Hinton2006}.\\

\noindent The time complexity of this process becomes forbiddingly large for big datasets and architectures. Hence there have been efforts at parallelization. These parallel algorithms fall in two broad categories \cite{Petrowski1993}. The algorithms of the first category partition the neural network and train the parts in parallel. The neural network can be physically partitioned \cite{Chen2012} or logically partitioned \cite{Deng2011, Deng2012}. The challenge is to ensure that the computational load is evenly distributed among the parallel processors and data communication is minimized. The second category of algorithms carry out calculations relative to the entire network but specific to subsets of data (single patterns or small batches of data) at different processing nodes \cite{DeGrazia2012}. This method is typically suitable for computer clusters as this requires very little communication. Each computing node $i$, generates its own update vector $\Delta w_i$ on the basis of the data-set $D_i$ assigned to it. At the end of an epoch when all the nodes are ready, the overall $\Delta w$ is calculated by averaging the outputs of each of the nodes.\\

\noindent As mentioned earlier, the pre-training phase is crucial for convergence but to the best of our knowledge, all existing methodologies perform pre-training layer-by-layer in a greedy way. In this work we propose a synchronized parallel layer-wise pre-training algorithm. The layers are learnt on parallel running threads that are synchronized regularly by cascading of maturity information. It achieves faster learning due to parallelism and reduces chances of misfit among the layers owing to synchronization.\\

\noindent Section \ref{sec:relevant_theory} briefly introduces some theoretical concepts relevant to stacked autoencoders. The synchronized layer-wise pre-training algorithm is described in Section \ref{sec:sync_pretrain}. Section \ref{sec:expt} presents an experimental methodology to compare the proposed algorithm with greedy layer-wise pre-training and Section \ref{sec:discussion} discusses the results of the experiment. Finally Section \ref{sec:conclusion} concludes the paper with a summary of the work.

\section{Preliminaries of stacked autoencoder}
In this section we describe stacked autoencoder and state briefly about the tools and methodologies used in their training.
\label{sec:relevant_theory}
	\subsection{Autoencoder}An \textit{autoencoder} is a type of neural network that is operated in unsupervised learning mode mainly for representation learning and dimensionality reduction \cite{Haykin1998}. It is aimed at encoding input $\textbf{x}$ into a representation $c(\textbf{x})$ so that the input can be reconstructed from the representation with minimum amount of deformation. Given a set of unlabelled training examples $\{\textbf{x}^{(1)}, \textbf{x}^{(2)}, ...\}$ the autoencoder sets the outputs $\textbf{y}^{(i)}=\textbf{x}^{(i)}$ and tries to learn a function $h_{W,b}$ such that $h_{W,b}(\textbf{x})=\hat{\textbf{x}}\approx \textbf{x}.$\\
	
\noindent A \textit{stacked autoencoder}\cite{Bengio2009,Hinton2006} has multiple hidden layers of neurons between the input and output layers. A stacked autoencoder of depth $n$ has an input layer, $2n-3$ hidden layers, and an output layer of neurons. Layers $1,2,...,n$ comprise the \textit{encoder} and the remaining layers form the \textit{decoder}. The $n^{th}$ layer is called the \textit{code layer}.\\

\noindent Autoencoder is usually trained by the backpropagation algorithm with an error criterion like mean squared error or cross entropy error \cite{Haykin1998}. 
	
	\subsection{Restricted Boltzmann Machine (RBM)}Restricted Boltzmann Machine (RBM) is commonly used for initializing deep neural networks like stacked autoencoder. An RBM is a Markov Random Field associated with an undirected graph that consists of m visible units $\textbf{V} = (V_1 , \dots,\allowbreak V_m)$ comprising the visible layer and $n$ hidden units $\textbf{H} = (H_1, \dots, H_n)$ comprising the hidden layer \cite{Bengio2009,Fischer2012}. There exists no connection among the units of the same layer in the graph. The joint probability distribution associated with the model is given by the Gibbs distribution:
	\begin{equation}
		p(\textbf{v,h}) = \frac{1}{Z} e^{-E(\textbf{v},\textbf{h})}
	\end{equation}
\noindent with the energy function:
	\begin{equation}
	E(\textbf{v,h})= -\displaystyle\sum_{i=1}^n \displaystyle\sum_{j=1}^m w_{i,j}h_iv_j - \displaystyle\sum_{j=1}^mb_jv_j - \displaystyle\sum_{i=1}^nc_ih_i
	\end{equation}		
The absence of connections among the nodes of the same layer implies that the hidden variables are conditionally independent given the state of the visible variables and vice versa:
	\begin{equation}
	p(\textbf{h}|\textbf{v})=\displaystyle\prod_{i=1}^n p(h_i|\textbf{v}) $$and$$ p(\textbf{v}|\textbf{h})=\displaystyle\prod_{i=1}^m p(v_i|\textbf{h})
	\end{equation}
We can use an RBM to model unknown probability distributions. The parameters are trained by gradient ascent on the log-likelihood of given data using Markov Chain  Monte Carlo methods like Contrastive Divergence \cite{Hinton2002}. 

	\subsection{Greedy Layer-wise pre-training}
\begin{figure}[!t]
\centering
\includegraphics[width = \textwidth]{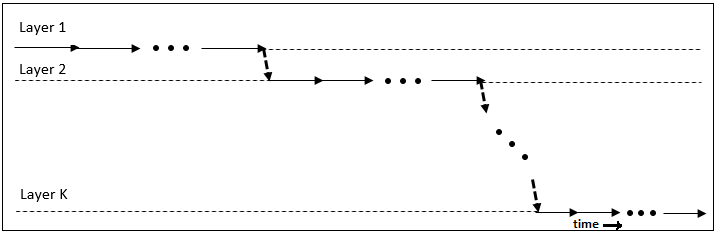}
\caption{Schematic of timing diagram for Greedy layer-wise pre-training. Each arrow indicates an epoch of learning while dashed line indicates idle time.} \label{fig:GLWP}
\end{figure}

Due to the presence of numerous local optima, proper initialization of the parameters is crucial for training deep neural networks. Greedy layer-wise pre-training was introduced with a motivation to roughly capture statistical abstractions of the input data in every layer and initialize the network parameters in a region near a good local optimum \cite{Bengio2006},\cite{Hinton2006}. Figure \ref{fig:GLWP} outlines the greedy layer-wise pre-training algorithm.\\

\noindent Let us consider a deep neural network with $K$ hidden layers of neurons. Let the input dimension be $N_i$, output dimension be $N_o$ and the hidden layer dimensions be $N_1,N_2,\dots,N_K$. Let us denote the training set by $S$. Greedy layer-wise pre-training follows one of the two methodologies described below.\\	 
	\paragraph{Pre-training using RBM:}
	In this approach, first an RBM with $N_i$ visible nodes and $N_1$ hidden nodes is trained. After training, the hidden weights and biases are used to initialize the incoming weights and biases respectively of the first hidden layer of the deep neural network. The input data of dimensionality $N_i$ is transformed to the dimension $N_1$ of the hidden layer. For an RBM with Bernoulli hidden units this is done by calculating the activation probabilities of the hidden units corresponding to the input examples $\textbf{v}\in S$ as follows:
	\begin{equation}
	p(H_i=1|\textbf{v})=\sigma\left(\displaystyle\sum_{j=1}^m w_{i,j}v_j+c_i\right)
	\end{equation}
	Now another RBM with $N_1$ visible units and $N_2$ hidden units is trained with the transformed data to find weights and biases which go and initialize the second hidden layer. The data is transformed to the dimension of the second hidden layer and this procedure is repeated for all of the hidden layers of the deep neural network.\\
	\paragraph{Pre-training using autoencoder:}
	Deep networks can also be initialized using autoencoders. In this methodology, single hidden-layer autoencoders are trained for each of the layers of the deep architecture and their weights and biases are used to initialize the corresponding layers of the deep architecture. 

\begin{figure}[!t]
\centering
\includegraphics[width = \textwidth]{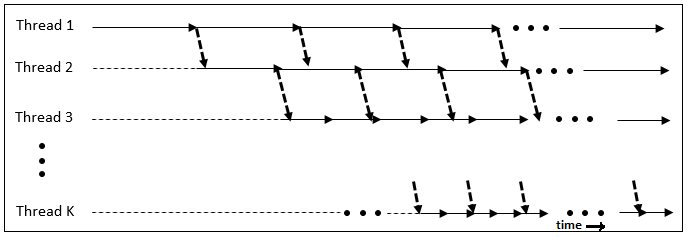}
\caption{Schematic of space-time diagram for proposed multi-core synchronized layer-wise pre-training. Like Figure \ref{fig:GLWP} each arrow indicates an epoch of learning and dashed lines mark the idle times of each layer.} \label{fig:SLWP}
\end{figure}

\section{Synchronized layer-wise pre-training}
\label{sec:sync_pretrain}
This section describes the proposed algorithm for accelerated training of stacked autoencoders. Greedy layer-wise pre-training makes a layer wait for all the previous layers to finish learning before it can start itself and for all the following layers to finish after it has finished. The guiding philosophy of the proposed algorithm is to reduce the idle time of greedy layer-wise pre-training by introducing parallelism with synchronization. This algorithm is suitable for multi-core machines where different layers can learn in parallel on threads running in different cores of the CPU.
\subsection{Methodology}
\noindent Let $L_l$ denote the $l^{th}$ layer of an autoencoder of depth $K$. The thread assigned to pre-train $L_l$ is denoted by $T_l$. The training and validation data for the $n^{th}$ epoch of its pre-training are denoted by $\textbf{D}^T_l[n]$ and $\textbf{D}^V_l[n]$ respectively. Let the activation function for $L_l$ be $f_l (.)$ and the incoming weights and biases after the $n^{th}$ epoch be $\textbf{W}^l[n]$ and $\textbf{b}^l[n]$ respectively $[\forall l = 1, 2,\dots ,K]$.\\

\noindent The training and validation data for the deep network are directly used for pre-training $L_1$ and correspond to $\textbf{D}^T_1[0]$ and $\textbf{D}^V_1[0]$ respectively. These are considered mature right from the beginning and are kept constant throughout. $\forall n>0$:
\begin{equation}
\textbf{D}^T_1[n] = \textbf{D}^T_1[0]\notag
\end{equation}
\begin{equation}
\textbf{D}^V_1[n] = \textbf{D}^V_1[0]
\end{equation}
As the training and validation data for $T_1$ are readily available, the algorithm starts with $T_1$ beginning to learn the first layer of the autoencoder. 
\begin{itemize}
	\item Due to data dependency $T_{l}$ waits until $T_{l-1}$ has completed one epoch of training. Once $T_l$ starts executing, every time it completes one epoch, it transforms $D^T_{l}$ and $D^V_{l}$ using the current weights and biases $\textbf{W}^l[n]$ and $\textbf{b}^l[n]$ and modifies $D^T_{l+1}$ and $D^V_{l+1}$ as:\begin{equation}
\textbf{D}^T_{l+1}[m+1] = f_l ( \textbf{W}^{l}[n] * \textbf{D}^T_{l}[n] + \textbf{b}^{l}[n] )\notag
\end{equation}
\begin{equation}
\textbf{D}^V_{l+1}[m+1] = f_l ( \textbf{W}^{l}[n] * \textbf{D}^V_{l}[n] + \textbf{b}^{l}[n] )
\end{equation}
$T_l$ executes for a stipulated $N_l$ epochs specified by the user. Cascading of knowledge synchronizes the threads and promotes harmonious training of the layers.

	\item After completing $N_l$ epochs, $T_l$ goes to sleep. If $T_{l-1}$ modifies $D^T_{l}$ and $D^V_{l}$ after that, $T_l$ wakes up, executes one iteration of learning with the modified data and goes back to sleep.

	\item Pre-training ends when all the threads have finished their stipulated number of iterations. 
	\item After pre-training has been completed, the entire architecture can be fine-tuned by backpropagation learning.
\end{itemize}

\noindent Pre-training can be done using RBM or autoencoder. Though described in terms of an autoencoder, this algorithm can also be used for pre-training other deep neural networks meant for classification, regression, etc.

\section{Experimental methodology}
\label{sec:expt}
The synchronized layer-wise pre-training algorithm was tested on the task of dimensionality reduction of \textit{MNIST handwritten digit dataset} using a stacked autoencoder. The reconstruction accuracy was measured using a mean squared error (MSE) crterion.
\subsection{Dataset}
MNIST\footnote{Available for free download at http://yann.lecun.com/exdb/mnist/} is a database of ($8$-bit) $28$x$28$ gray-scale handwritten digits, popularly used for training and testing image processing and machine learning systems. The database contains $60,000$ training examples and $10,000$ test examples.\\
\noindent For our experiments, the training set was further divided into $50,000$ training examples and $10,000$ validation examples. The validation set contained $1000$ examples from each class of digits $(0-9)$ which ensures equal participation from all of the classes. The training set was randomized and divided into minibatches of size $100$ before every new epoch of pre-training and fine-tuning.\\
\noindent Figure \ref{fig:serialOriginal} shows some sample digits from the dataset.

\subsection{Parameters of the stacked autoencoder}
Details of the architecture of the stacked autoencoder have been presented in Table \ref{table:archSpec}. This is same as the architcture used by \cite{Hinton2006} for the same task with the only exception of sigmoid activation functions for all the layers.
\begin{table}[!h]
\begin{center}
\caption{Architectural parameters of the stacked autoencoder}
\begin{tabular}{|c|c|}
\hline
\textbf{Parameter}      &   \textbf{Value}\\ \hline 
         Depth          &       $5$         \\ \hline
    Layer dimensions    &   $784, 1000, 500, 250, 30$\\ \hline
   Activation function  &     $sigmoid$     \\ \hline
\end{tabular}
\label{table:archSpec}
\end{center}
\end{table}
\vspace{-20pt}
\subsection{Parameters of the learning algorithms}
\noindent The learning rate of both the weights and biases was set at $0.1$ for contrastive divergence and $0.001$ for backpropagation. In addition a momentum of $0.5$ for the first $5$ epochs and $0.9$ for the remaining epochs was used in contrastive divergence learning.\\

\noindent The performances of greedy layer-wise pre-training (the baseline) \cite{Hinton2006} and the proposed synchronized layer-wise pre-training algorithm were compared on a system with $8$ CPU cores (dual-hyperthreaded quad-core) and $8GB$ RAM in terms of reconstruction error calculated as the average squared reconstruction error per digit and the total training time. The baseline experiment was performed with $20$ epochs of RBM pre-training for every layer followed by $10$ epochs of fine-tuning of the entire architecture via backpropagation. Next, the proposed algorithm was executed for the same amount of pre-training and fine-tuning but the threads $L_2$, $L_3$ and $L_4$ carried out $5, 20$ and $40$ additional epochs of pre-training respectively, every time the previous thread modified their data.\\

\begin{figure*}[!h]
\centering
\includegraphics[width = \textwidth]{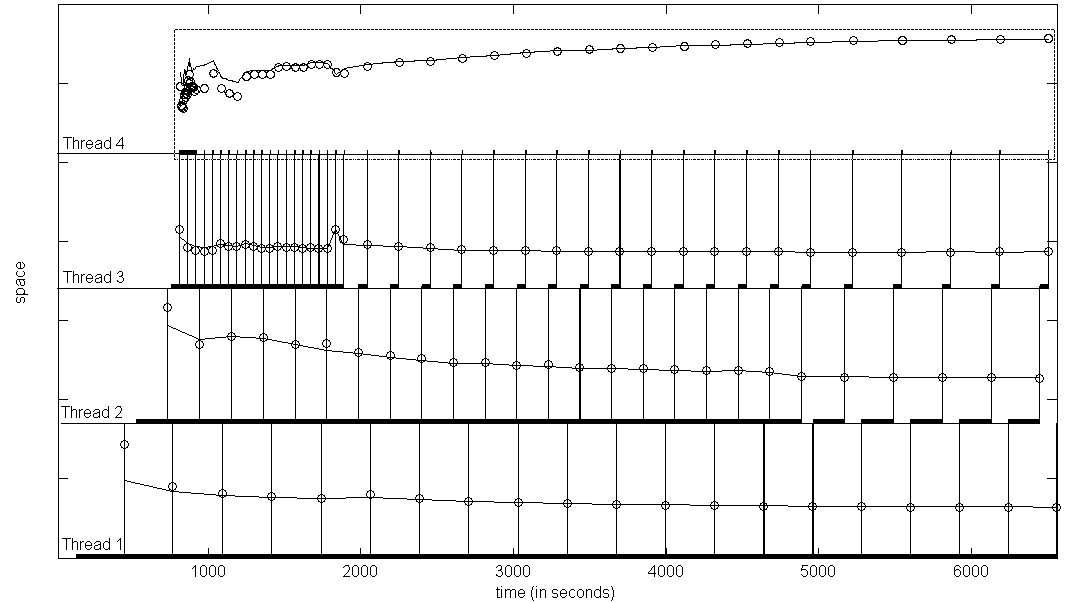}
\caption{Space-time diagram of the proposed synchronized layer-wise pre-training algorithm. The compute times of each of the threads are marked with thick lines. The black spikes between the threads indicate modification of data of the upper thread by the lower thread after every epoch of pre-training. Plots of the reconstruction errors for training (solid line) and validation data (circles) for each of the RBMs have been shown above the timing diagrams for the corresponding threads.}

\label{fig:threadInteraction}
\end{figure*}

\begin{figure*}[!h]
	\centering
	\includegraphics[width = 0.9\textwidth]{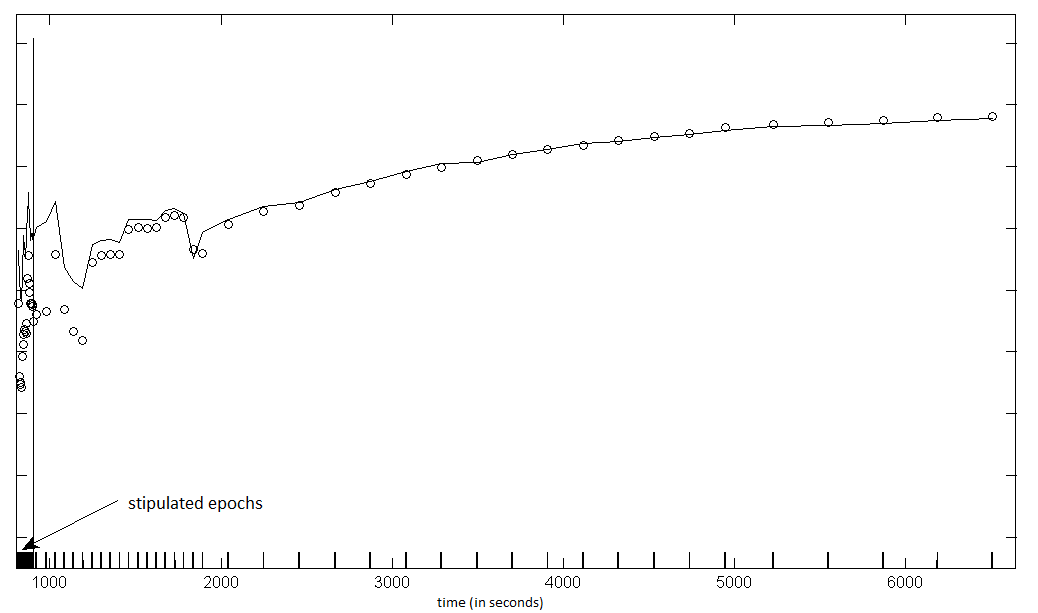}
	\caption{Magnified view of the timing diagram for the innermost layer (marked by dotted rectangle in Figure \ref{fig:threadInteraction}). Notice how $L_4$ gradually corrects its initial over-fitting as the training data matures. The solid line indicates training error and the circles denote validation error}
	\label{fig:AE4zoom}
\end{figure*}

\section{Results and discussion}
\label{sec:discussion}
This section presents the results of the comparison and discusses a few aspects of the proposed algorithm.
\subsection{Convergence} 
\label{convergence}
\begin{figure*}[!h]
\centering    
\subfigure[]{\label{fig:serialOriginal}\includegraphics[width=0.25\textwidth]{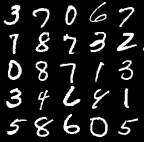}} 
\subfigure[]{\label{fig:serialRecons}\includegraphics[width=0.25\textwidth]{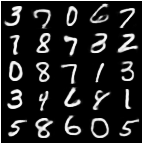}}
\subfigure[]{\label{fig:parallelRecons}\includegraphics[width=0.25\textwidth]{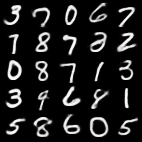}}
\caption{Comparison of performance: (a) Original (b) reconstruction with greedy layer-wise pre-training (c) reconstruction with synchronized layer-wise pre-training}
\label{fig:Recons}
\end{figure*}

\begin{figure*}
\centering
\includegraphics[width = \textwidth]{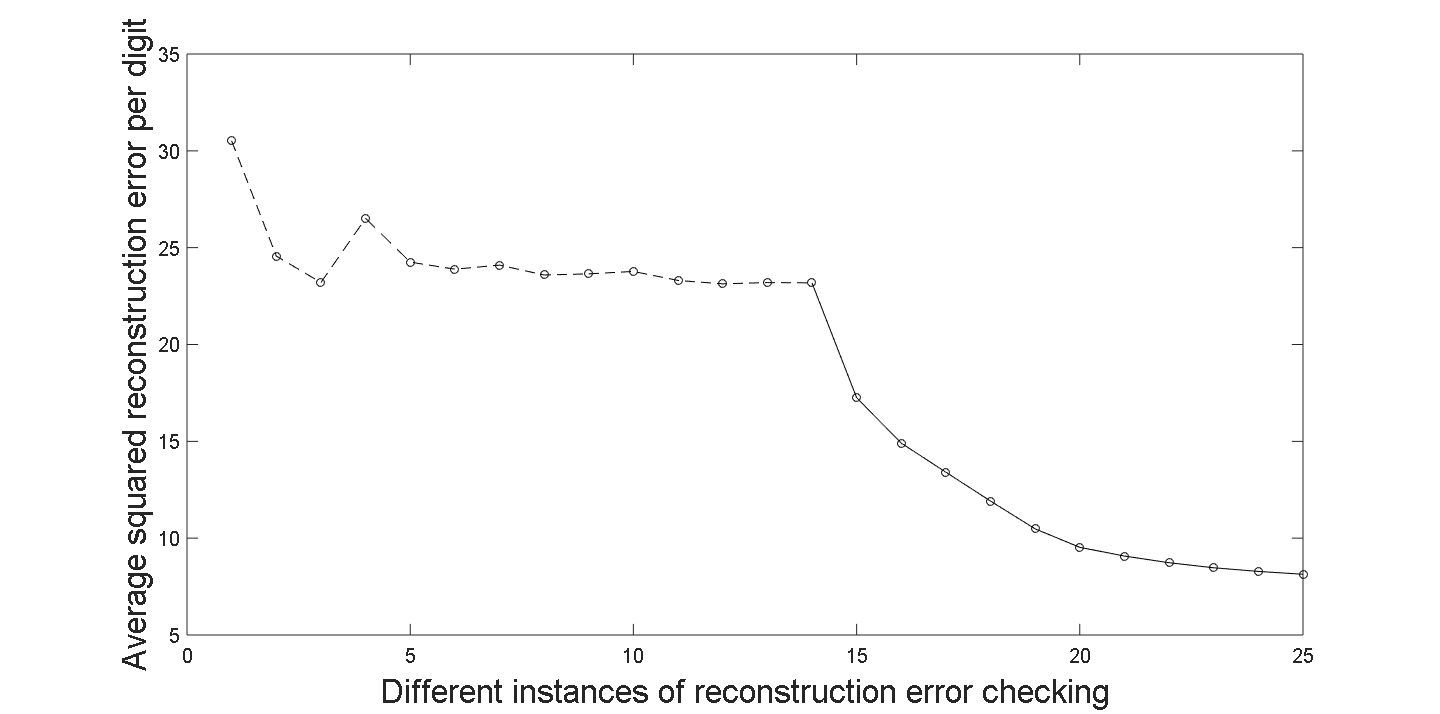}
\caption{Variation of the error in reconstruction of validation set for the stacked autoencoder during synchronized pre-training (dashed line) and fine-tuning (solid line). Reconstruction error was checked at regular intervals during pre-training and after every epoch of backpropagation during fine-tuning}
\label{fig:ErrVar}
\end{figure*}

Figure \ref{fig:ErrVar} shows the variation of reconstruction error of the validation set during the execution of the proposed algorithm. Table \ref{table:ReconsErr} gives the final reconstruction errors for the two experiments. Figure \ref{fig:Recons} compares the reconstructions of 25 randomly chosen samples from the test set by the two methods. The performances are almost equivalent.\\

\begin{table}[]
\begin{center}
\caption{Average squared reconstruction error per digit}
\begin{tabular}{|p{2cm}|c|c|}
\hline
\textbf{Algorithm} & \textbf{Training Error} & \textbf{Test Error}\\ \hline
       Greedy pre-training &   8.00  &  8.19\\ \hline
       Synchronized pre-training &   8.39  &  8.57\\ \hline
\end{tabular}
\label{table:ReconsErr}
\end{center}
\end{table}

\begin{table}[]
\begin{center}
\caption{Execution times}
\begin{tabular}{|p{2cm}|c|c|}
\hline
\textbf{Algorithm} & \textbf{Pre-training time} & \textbf{Fine-tuning time}\\ \hline
       Greedy pre-training &   3h 14m 43s  &  2h 16m 59s\\ \hline
       Synchronized pre-training &   1h 49m 11s  &  2h 15m 42s\\ \hline
\end{tabular}
\end{center}
\label{table:time}
\end{table}


\subsection{Speed-up}
Table \ref{table:time} compares the execution times of the two algorithms. The proposed algorithm achieves the same performance $1h 26m 49s$ faster than greedy layer-wise pre-training which is a $26.17\%$ speedup. The four threads, $T_1$ through $T_4$ were run on four different cores of the same CPU. The time for communication of data among the threads was observed to be $5$ to $6$ orders of magnitude lesser than the time for an epoch of RBM pre-training. \\

\noindent In our experiment pre-training of the stacked autoencoder was concluded as soon as the RBM for the first layer ($784-1000$) completed its stipulated 20 epochs. Though it changed $D^T_2$ and $D^V_2$ after the $20^{th}$ epoch, the parameters of $L_2$ or any of its data-dependent layers were not updated. This was just to keep the total pre-training time equal to the time taken for $L_1$ (the largest and the slowest to train layer) to complete its prescribed amount of pre-training. Performing an extra round of update for $L_2, L_3$ and $L_4$ prolonged the training but did not improve the final reconstruction accuracy.\\

\subsection{Analysis of the space-time diagram}
Figure \ref{fig:threadInteraction} shows the space-time diagram for the proposed algorithm. The variation of reconstruction errors of the training and validation examples have been plotted for each layer. Both the errors show a decreasing trend for the first three layers indicating the parameters of these layers, getting tuned to the given data. Initially the training and validation errors differ for all the layers but eventually converge to almost equal values. This indicates that the parameters achieve generalization beyond the training set in the course of pre-training. \\

\noindent The behavior of the fourth layer is particularly interesting. Figure \ref{fig:AE4zoom} zooms into its timing diagram (marked with a dotted rectangle in Figure \ref{fig:threadInteraction}). The validation and training errors differ by a large margin for quite sometime (throughout the stipulated amount of pre-training and beyond) but eventually converge as $T_4$ updates its parameters every time $D^T_4$ and $D^V_4$ are changed by $T_3$. Also both the training and validation errors show an increasing trend throughout the process. Low initial reconstruction error that increases with the progress of training indicates that the random values to which the weights and biases of $L_4$ were initialized, overfit the data that was provided by $L_3$ initially. Significant difference between the training and validation errors during this period of time further corroborates the overfitting. However with the progress of training, as the data matures, the overfitting slowly cures and the training and validation errors converge to close values. The increasing trend of the reconstruction error implicitly indicates how synchronization forces the innermost layer to increase its own error to conform to the remaining of the network. The behavior of the fourth layer also indicates the necessity of updating the parameters even after the stipulated amount of pre-training has been completed. As the training data of a hidden layer changes throughout the process of pre-training, the parameters must also be updated regularly to keep up with the changing data. This does not add an overhead to the total pre-training time which is always dictated by the time taken by the largest (or the most complex) layer to complete the stipulated number of epochs of pre-training. 

\section{Conclusion}
\label{sec:conclusion}
In this work the existing methods of parallelizing deep architecture learning have been explored and a new algorithm for parallelizing the pre-training phase has been proposed. The convergence and performance of the algorithm have been tested on the task of dimensionality reduction of MNIST handwritten digit database using stacked autoencoder and 26\% speed-up has been observed for achieving the same performance. The key feature of the proposed algorithm is that it is not greedy. This reduces chances of misfit among the layers, a common hazard of greedy learning and minimizes the idle time for each layer. The proposed algorithm can be seamlessly integrated with the existing methods of parallelization which mainly concentrate on the fine-tuning phase.
\bibliographystyle{IEEEtran}
\bibliography{BTP_citations}
\end{document}